\documentclass[sigconf,authordraft]{acmart}

\usepackage{multirow}
\usepackage{graphicx}
\usepackage{booktabs} 
\usepackage{dsfont}

\AtBeginDocument{%
  \providecommand\BibTeX{{%
    \normalfont B\kern-0.5em{\scshape i\kern-0.25em b}\kern-0.8em\TeX}}}

\setcopyright{none}
\renewcommand\footnotetextcopyrightpermission[1]{}
\copyrightyear{2021}
\acmYear{2021}
\acmDOI{10.1145/1122445.1122456}
\begin{document}
\title{\emph{NN-LUT}: Neural Approximation of Non-Linear Operations for Efficient Transformer Inference}

 \author{Joonsang Yu}
 \affiliation{%
   \institution{NAVER AI Lab}
   \institution{Face, NAVER Clova}
   \country{}
 }
 \authornotemark[1]
 \email{joonsang.yu@navercorp.com}
 
 \author{Junki Park}
 \affiliation{%
   \institution{SAIT}
   \country{}}
 \email{junki.park@samsung.com}
 
  \author{Seongmin Park}
  \affiliation{%
   \institution{Hanyang University}
   \country{}}
  \email{skstjdals@hanyang.ac.kr}
 
  \author{Minsoo Kim}
  \affiliation{%
   \institution{Hanyang University}
   \country{}}
  \email{minsoo2333@hanyang.ac.kr }
 
  \author{Sihwa Lee}
  \affiliation{%
   \institution{Hanyang University}
   \country{}}
  \email{macto94@hanyang.ac.kr}

  \author{Dong Hyun Lee}
  \affiliation{%
  \country{}
   }
  \authornote{Works done at SAIT.}
  \email{starcafe1224@gmail.com}
 
  \author{Jungwook Choi}
  \affiliation{%
   \institution{Hanyang University}
   \country{}}
  \email{choij@hanyang.ac.kr}
 







\begin{abstract}
Non-linear operations such as GELU, Layer normalization, and Soft-max are essential yet costly building blocks of Transformer models. Several prior works simplified these operations with look-up tables or integer computations, but such approximations suffer inferior accuracy or considerable hardware cost with long latency. This paper proposes an accurate and hardware-friendly approximation framework for efficient Transformer inference. Our framework employs a simple neural network as a universal approximator with its structure equivalently transformed into a LUT. The proposed framework called \emph{NN-LUT} can accurately replace all the non-linear operations in popular BERT models with significant reductions in area, power consumption, and latency.

\end{abstract}





\maketitle

\section{Introduction}
\label{sec:introduction}

The Transformer-based pre-trained neural networks such as BERT~\cite{devlin2019bert} and RoBERTa~\cite{liu2019roberta} have achieved significant success in improving the performance of various Natural Language Processing (NLP) tasks. These models are characterized by the self-attention mechanism, which links different symbols within a sequence to obtain a relational representation.
Thanks to the exceptional performance of the pre-trained Transformer models, there has been an increasing need for their efficient deployment. However, Transformer's computational characteristics hinder straightforward implementation. The most acknowledged issue is the gigantic size of the pre-trained Transformer models; for example, GPT-3~\cite{floridi2020gpt} contains 175 billion parameters with 2048 tokens, incurring profound memory and computation overhead. Seminal research efforts attempted to reduce this burden; \cite{bhandare2019efficient,zafrir2019q8bert} quantized bit-precision to reduce memory footprint and expedite BERT inference. \cite{wang2020spatten} further proposed dynamic pruning and quantization to decrease computational complexity on the fly. Although these efforts have proved significant potential in alleviating Transformer's computational burden, their focus has been limited to matrix multiplication.    

Another important yet less examined characteristic of the pre-trained Transformer is its mixture of non-linear computations for realizing the self-attention mechanism. A typical Transformer computation involves several non-linear operations such as GELU activation function, Softmax, and Layer normalization (LayerNorm). They are generally implemented using high-precision floating-point computations, requiring expensive, slow 32-bit floating-point (FP32) arithmetic units. Since these non-linear operations are embedded within Transformer's basic computation block, their inefficiency becomes a significant slow-down factor for overall Transformer computation as reported by \cite{stevens2021softermax, shkim2021ibert}.

Several prior works attempted to address the computational bottleneck of non-linear operations. \cite{shkim2021ibert} proposed approximation techniques to compute GELU, Softmax, and LayerNorm using only 32-bit integer (INT32) arithmetic. \cite{stevens2021softermax} also proposed hardware-friendly reduced-precision computation techniques and a custom accelerator for Softmax. These state-of-the-art techniques have demonstrated successful approximation of non-linear operations with negligible accuracy degradation. However, these approaches are based on operation-specific multi-step computations with complicated data paths and increased latency, hindering the practical implementation to the existing neural processing units (NPUs). Also, \cite{shkim2021ibert} and \cite{stevens2021softermax} take advantage of "approximation-aware fine-tuning" to adjust the entire model parameters for compensation of approximation errors. Such fine-tuning requires expensive training computation and labeled datasets, thus prohibiting the off-the-shelf fine-tuned models.  

In this paper, we propose a general approximation framework called \emph{NN-LUT} for non-linear operations of pre-trained Transformer models. \emph{NN-LUT} employs a neural network as a universal approximator (\cite{esmaeilzadeh2012neural, eldridge2014neural, peng2018axnet}). But it is structured as a one-hidden-layer ReLU network to transform it into an equivalent yet hardware-friendly LUT approximation. Therefore, the same \emph{NN-LUT} hardware can approximate various non-linear operations by simply updating the LUT contents, particularly attractive for NPUs already equipped with LUTs \cite{NVDLA,song20197,jang2021sparsity}. We further propose three techniques to enhance \emph{NN-LUT}'s accuracy: 1)  a generally applicable strategy for \emph{NN-LUT} training, 2) input scaling for a wide-range approximation, and 3) a dataset-free lightweight \emph{NN-LUT} calibration method. We demonstrate that 
\emph{NN-LUT} can accurately replace all the non-linear operations of RoBERTa~\cite{liu2019roberta} and MobileBERT~\cite{sun2020mobilebert} with negligible accuracy degradation on GLUE and SQuAD tasks. 
From hardware cost analysis based on hardware synthesis with commercial 7-nm technology, we demonstrate that \emph{NN-LUT} achieves $2.63\times$, $36.4\times$, and $5.9 - 9.8\times$ savings in area,  power consumption, and latency, respectively, compared to the hardware for the state-of-the-art integer approximation of \cite{shkim2021ibert}. We further integrate the custom hardware into an existing neural processing unit (\cite{jang2021sparsity}) and conduct system-level performance analysis, showcasing up to 26\% system speedup solely thanks to \emph{NN-LUT}'s hardware efficient approximation of non-linear operations. 
Therefore, we conclude that \emph{NN-LUT} is an attractive approximation framework for the efficient deployment of pre-trained Transformer models.  

\begin{itemize}
    \item We propose a novel transformation of one-hidden-layer ReLU neural network into LUT-based approximation for hardware-friendly implementation, called \emph{NN-LUT}. 
    \item To improve \emph{NN-LUT}'s accuracy, we suggest a general strategy for \emph{NN-LUT} training, input scaling for a wide-range approximation, and dataset-free \emph{NN-LUT} calibration.
    \item We demonstrate that \emph{NN-LUT} works as a drop-in replacement of GELU, Softmax, and LayerNorm with little accuracy degradation for popular BERT models and tasks.
     \item We synthesize arithmetic units with the 7-nm technology and analyze system-level performance that highlights the superiority of \emph{NN-LUT} compared to the state-of-the-art integer approximation.
\end{itemize}

\section{Preliminaries and Related Work}
\label{sec:relatedwork}

\subsection{Non-Linear Operations of Transformers}
A typical Transformer architecture consists of a multi-head attention block followed by a feed-forward block \cite{vaswani2017attention}. Several non-linear operations are involved in the Transformer computation: Softmax for extracting self-attention features, GELU for activation within the feed-forward block, LayerNorm for normalization of each block output. GELU, Softmax, and LayerNorm are defined as follows:



\begin{equation}
\begin{aligned}
\label{eq:gelu}
\text{GELU}(x) := \frac{x}{2}\big[1+\text{erf}(\frac{x}{\sqrt{2}})\big]
\text{; erf}(x) := \frac{2}{\sqrt{\pi}}\int_{0}^{x} \text{exp}(-t^2) \,dt
\end{aligned}
\end{equation}

\begin{equation}
\begin{aligned}
\label{eq:softmax}
\text{Softmax}(x_i) := \frac{\text{exp}(x_i)}{\sum_{j=1}^{k}\text{exp}(x_j)}
\end{aligned}
\end{equation}

\begin{equation}
\begin{aligned}
\label{eq:layernorm}
\text{LayerNorm}(x_i) := \frac{x_i-\mu}{\sigma}\text{; }\mu=\frac{1}{C}\sum_{i=1}^{C}x_i \text{, } \sigma=\sqrt{\frac{1}{C}\sum_{i=1}^{C}(x_i-\mu)^2}
\end{aligned}
\end{equation}

As shown in the equations, these non-linear operations take significant run-time due to costly FP32 arithmetic computation \cite{stevens2021softermax}. Therefore, numerous approximation techniques have been proposed to alleviate the hardware costs.  
 
\subsection{Neural Network Approximation} 

The neural networks are considered as universal function approximators \cite{cybenko1989approximation, hornik1991approximation}. Motivated by approximation capability of neural networks, \cite{esmaeilzadeh2012neural} proposed an acceleration framework that approximated compute-intensive code regions (e.g., FFT) with multi-layer perceptrons. \cite{eldridge2014neural} also employed a simple neural network to approximate costly transcendental functions. Furthermore, neural network structures have been evolved into approximators and predictors to improve approximation accuracy \cite{peng2018axnet}. However, the strong approximation capability of neural networks comes at the cost of a large number of matrix computations. To address this issue, we propose a method to transform the neural network computation into a simple LUT operation (i.e., a table look-up + one MAC), significantly reducing the computational overhead of neural network approximation. This computation-efficient approximation is essential for accelerating non-linear operations of Transformers, which take a significant portion of the overall computational load.   

\subsection{Implementation of Non-Linear Operations} 
Implementing arithmetic calculation of non-linear operations is challenging since they involve transcendental functions like exponential. A conventional approach is to devise a deeply pipelined arithmetic unit to increase computation throughput \cite{chen2017high}. However, this arithmetic unit consists of complex datapaths and controls for handling floating-point computation, incurring long latency and significant hardware overhead. 
The recent studies \cite{shkim2021ibert,stevens2021softermax} approximate non-linear functions with reduced-precision arithmetic computation. However, these approaches create different computation sequences for each non-linear operation, hindering a unique hardware implementation applicable to all the non-linear operations. Unlike these previous works, \emph{NN-LUT} is a general neural-net-based approximation framework implemented with a single LUT. Therefore, the area/resource overhead of \emph{NN-LUT} does not grow no matter how many non-linear operations it targets.

\section{Method}
\label{sec:method}






\subsection{Look-Up Table Approximation}
Look-up table (LUT) approximation is a hardware-efficient technique, equipped in various neural network accelerators \cite{NVDLA,park20219}. It is constructed as an $N$-entry table containing piece-wise approximation parameters on a target input range. Thanks to its simple hardware implementation, a first-order approximation (i.e., $y=sx+t$) is popular; it takes only one multiplier and adder for output computation. Given an $N$-entry table of the approximation parameters $\{s_i, t_i\}_{i=1:N}$ and the breakpoints $\{d_i\}_{i=1:N-1}$, LUT$(x)$ can be defined as follows:

\begin{equation}
\begin{split}
\label{eq:lut}
\text{LUT}(x) := 
    \begin{cases}
        s_1x + t_1 &\text{ if } x < d_1\\
        s_ix + t_i &\text{ if } d_{i-1} \leq x < d_i \;(\text{for } 1 < i \leq N-1)\\
        s_Nx + t_N &\text{ if } x \geq d_{N-1}.
    \end{cases}       
\end{split}
\end{equation}

The approximation parameters of LUT are chosen to best approximate a target function $f(x)$. Popular methods include curve fitting and interpolation ~\cite{cantoni1971optimal}. The breakpoints determine the interval of the approximation range. Linear-mode divides the input range into equally-spaced breakpoint intervals, whereas Exponential-mode has shorter intervals on low range values and longer intervals on high range values \cite{NVDLA}. This pre-determined breakpoint allocation simplifies LUT hardware, but it imposes constraints on the location of breakpoints, negatively affecting the approximation accuracy.  

\begin{figure*}[!t]
\centering
\includegraphics[width=0.8\linewidth]{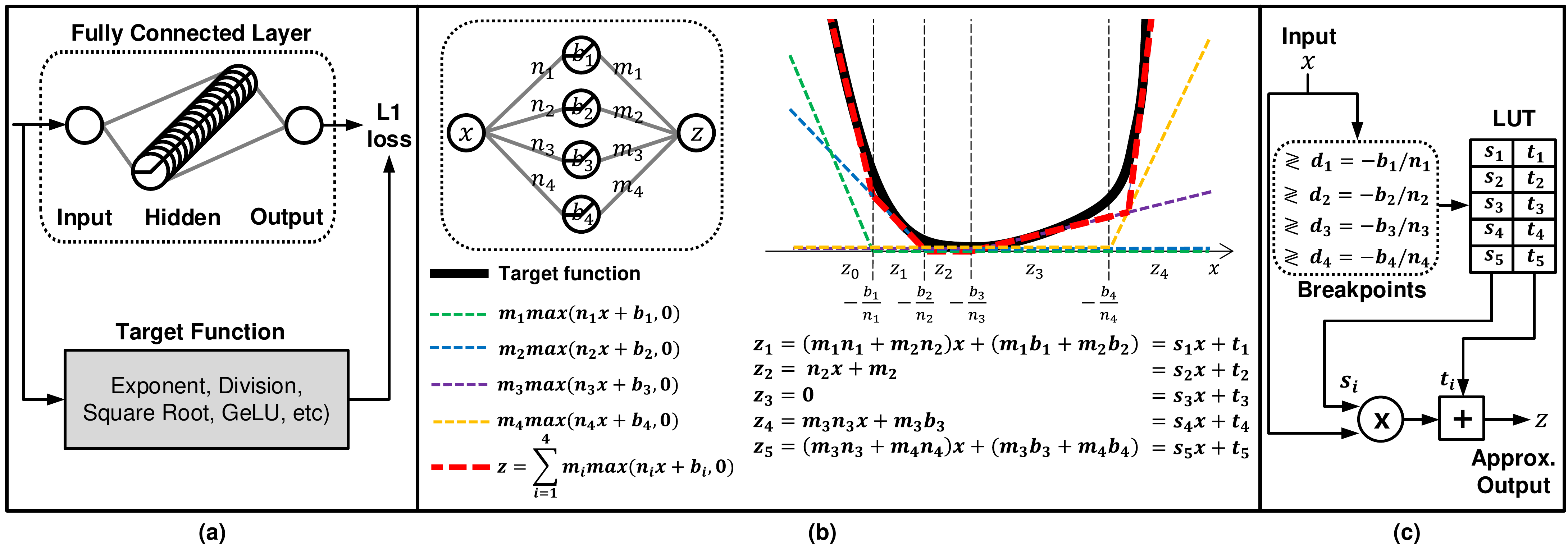}
\vspace{-5mm}
\caption{\emph{NN-LUT} overview: (a) 1-hidden-layer ReLU neural network for approximation of non-linear operations, (b) Neural network to look-up table conversion, (c) LUT implementation (best shown in color).}
\label{fig_nn-lut}
\vspace{-3mm}
\end{figure*}

\subsection{LUT-based Neural Network Approximation}
\label{sec:nn-lut}
Determining breakpoints that minimize the approximation error is not trivial. We approach this problem from a novel perspective of neural network approximation of non-linear operations. In particular, we claim that a one-hidden layer ReLU network, which is known to be a universal approximator (\cite{cybenko1989approximation,lu2017expressive}), can be transformed into an equivalent look-up table of Eq.~\ref{eq:lut}. 

Let NN$(x)$ be a one-hidden layer neural network of $N-1$ neurons with ReLU activation $\sigma$ parameterized with the first layer weight $n_i$ and bias $b_i$, and the second layer weight $m_i$: 

\begin{equation}
\label{eq:NN}
\text{NN}(x) = \sum_{i=1}^{N-1}m_i \sigma(n_ix+b_i) = \sum_{i=1}^{N-1} m_i y_i. 
\end{equation}

Note that $y_i$ is a hidden neuron obtaining a non-zero value only for a certain condition on the parameters $n_i, b_i$ and the input $x$. Without loss of generality, assume that a set of breakpoints, $\{-\frac{b_i}{n_i}\}_{i=1:N-1}$, is sorted in an ascending order that covers the input range. For each interval $-\frac{b_i}{n_i} < x < -\frac{b_{i+1}}{n_{i+1}}$, check the value of all the hidden neurons $\{y_j\}_{j=1:N-1}$ as follows:
\begin{equation}
\begin{split}
\text{For } j\leq i \text{, }\; y_j & =
    \begin{cases}
       n_j x + b_j &\text{ if } n_j \geq 0\\
       0 &\text{ otherwise}
    \end{cases}\\
\text{For } j > i \text{, }\; y_j & =
    \begin{cases}
       n_j x + b_j &\text{ if } n_{j+1} < 0\\
       0 &\text{ otherwise}
    \end{cases}/
\end{split}
\end{equation}

In other words, hidden neurons $y_j$ on the left of the current interval are non-zero if and only if their scale $n_j$ is positive, and the opposite is true for the hidden neurons on the right. Therefore, one can decompose NN$(x)$ into a linear function $z_i(x)$ defined on each interval ($n_j^+ = n_j \cdot (\mathds{1})(n_j\geq0)$ and $n_j^- = n_j \cdot (\mathds{1})(n_j<0)$):
\begin{equation}
\begin{split}
\label{eq:nn-lut}
 z_i &= \sum_{j=1}^{N-1} m_j y_j \;\text{ if } (-\frac{b_i}{n_i} < x < -\frac{b_{i+1}}{n_{i+1}}) \\
     &= m_i \Big[\sum_{j=1}^{i} n_j^+ + \sum_{j=i+1}^{N-1} n_j^-\Big] x + m_i \Big[\sum_{j=1}^{i} n_j^+ \frac{b_j}{n_j} + \sum_{j=i+1}^{N-1} n_j^- \frac{b_j}{n_j}\Big] \\
     &= s_i x + t_i \;\text{ if } (d_i < x < d_{i+1}),
\end{split}
\end{equation}

Therefore, a neural network, NN($x; \{m_i, n_i, b_i\}_{i=1:N-1}$), is transformed into a look-up table operation with parameters $\{s_i, t_i\}_{i=1:N}$ and breakpoints $\{d_i\}_{i=1:N-1}$).

Figure~\ref{fig_nn-lut} illustrates overall procedure of \emph{NN-LUT}. A 1-hidden-layer ReLU neural network is first trained offline with the target non-linear function for approximation. As shown in Figure~\ref{fig_nn-lut}(b), the trained neural network has hidden neurons described in different colors, which compose an approximation (the red piece-wise linear line) of the target function (the black line). Decomposition of this approximation into breakpoints $d_i$ results in a set of linear functions $z_i$ defined by approximation parameters $s_i, t_i$ for each interval $z_i$. As shown in Figure~\ref{fig_nn-lut}(c), the obtained approximation parameters and breakpoints can be implemented as a LUT-based approximation.

Note that the LUT parameters and the breakpoints in Eq.~(\ref{eq:nn-lut}) are constant values once the approximation network is trained for a target non-linear operation. Therefore, this LUT-based neural network approximator (thus called \emph{NN-LUT}) can take advantage of both the approximation capability of the neural network and the hardware-friendly LUT-based deployment.

\subsection{Improving \emph{NN-LUT} Performance}
\label{sec:strategy}
Despite the approximation capability of neural networks, care is needed to approximate the target operation with limited LUT entries successfully. 
This section presents several practical strategies for training \emph{NN-LUT} for approximating non-linear operations.

\subsubsection{Training Setup for \emph{NN-LUT}}
\label{sec:strategy-param-init}
The training dataset of \emph{NN-LUT} can be automatically generated by entering a set of input data into the target non-linear operations. Given an input range of interest, we uniformly sample values within the range. Since non-linear operations have different domains of interest, it is crucial to decide a proper range of input data for the successful training of approximation networks. Furthermore, the neural network parameters ($m_i$,$n_i$,$b_i$) should be properly initialized to well find the LUT parameters ($s_i$,$t_i$,$d_i$). 
Table~\ref{tab:nn-train} summarizes the strategy of input data range and the parameter initialization for \emph{NN-LUT} approximation of each non-linear operation.


\begin{table}[]
\caption{Training Setup for \emph{NN-LUT}.}
\vspace{-3mm}
\label{tab:nn-train}
\centering
\resizebox{\linewidth}{!}{%
\begin{tabular}{|c|c|c|c|c|}
\hline
Non-Linear Ops           & Function & Input Data  & Weight Init ($n_i$)    & Bias Init ($b_i$)      \\ \hline
GeLU                     & GeLU     & (-5, 5)     & Random          & Random          \\ \hline
\multirow{2}{*}{Softmax} & Exp      & (-256, 0)   & Positive Random & Positive Random \\ \cline{2-5} 
                         & Divide   & (1,1024)    & Negative Random & Positive Random \\ \hline
LayerNorm                & 1/SQRT   & (0.1, 1024) & Negative Random & Positive Random \\ \hline
\end{tabular}%
}
\vspace{-4mm}
\end{table}

\subsubsection{Input Scaling for a Wide Range Approximation}

It is observed that 1/SQRT of LayerNorm has a broad output dynamic range when input is smaller than one. If the output activation of a self-attention layer has a small variance ($\ll 1$), it results in a large output of 1/SQRT. This functional behavior is not desirable for a neural network approximator; neural network parameters need to be drastically adjusted to make steep slopes. To alleviate this issue, we propose an input scaling method: 1) Learn the model for the input range of $1$ to $K$ ($K \gg 1$) to be well trained over this wide range of monotonous output. 2) If small input ($0 < x < 1$) enters, it is mapped to the range of $1$ to $K$ by multiplying a large constant scale, $S$, then multiply the corresponding output from LUT by $\sqrt{S}$. 
One can implement the proposed method effectively by choosing $S$ as a power-of-two number, e.g., $2^{10}$; thus, a scaling operation becomes a simple bit-shift. In Sec.~\ref{sec:sw-eval}, we empirically show that this simple two-step scaling method can achieve high accuracy approximation when applied in BERT inference. 

\subsubsection{Calibration of \emph{NN-LUT} parameters}
One unique advantage of \emph{NN-LUT} is that the LUT parameters can be calibrated to approximate the non-linear operations better. This feature is handy for Transformer models with varying dynamic ranges across the layers. Commonly, each non-linear operation of a down-streamed Transformer model is replaced by the \emph{NN-LUT} trained offline, called direct approximation. If direct approximation's accuracy loss is noticeable, one can run \emph{NN-LUT} calibration with a small set of unlabeled data; each \emph{NN-LUT} is regressed with its full-precision reference function to update the approximation network parameters. Since all the Transformer parameters are frozen, the calibration can be promptly finished (less than 5\% of a typical fine-tuning time). Once calibration is done, the NN parameters are transformed into the LUT parameters (following Eq.~\ref{eq:nn-lut}) for efficient inference.

\begin{figure}[!t]
\centering
\includegraphics[width=\linewidth]{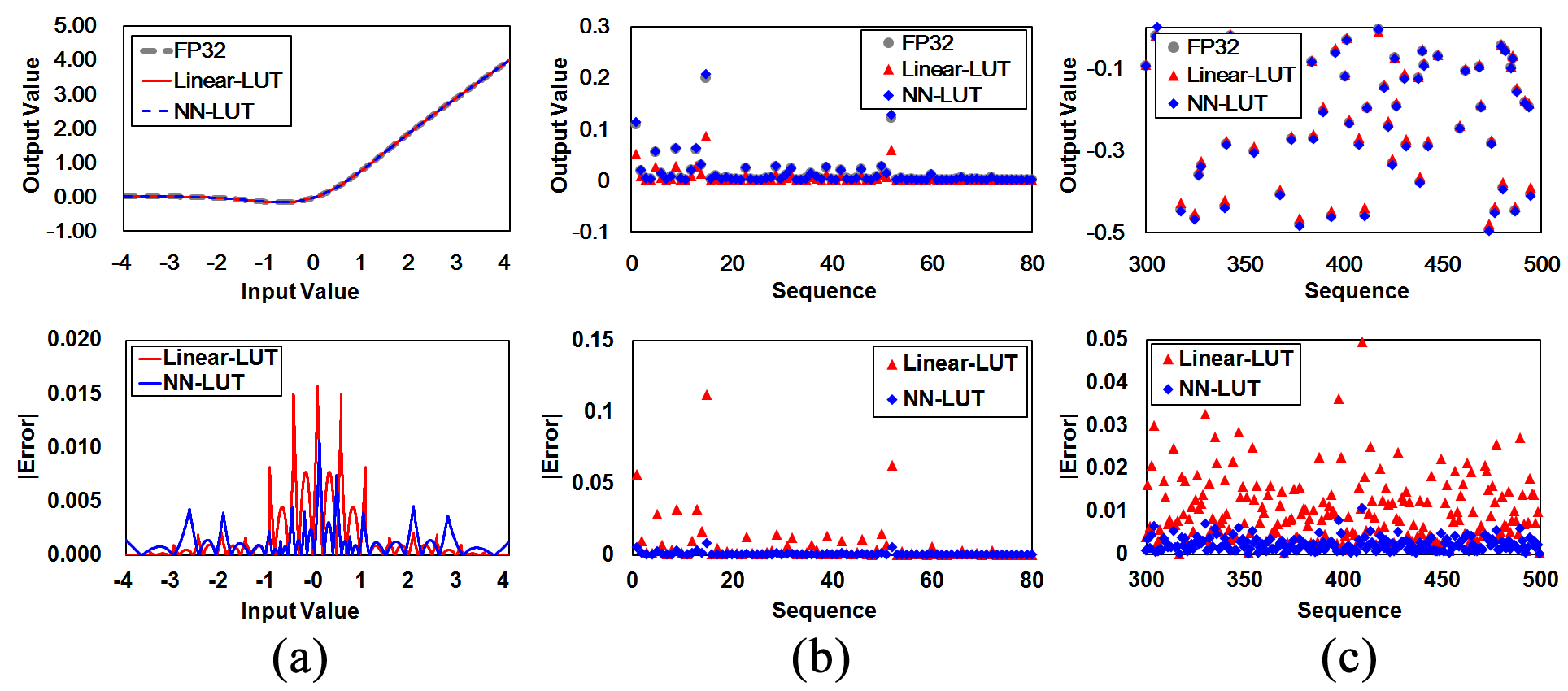}
\vspace{-7mm}
\caption{Approximation accuracy for non-linear operators: (a) GELU, (b) Softmax, (c) LayerNorm.}
\vspace{-3mm}
\label{fig:expr-non-linear-ops}
\end{figure}

\section{Software Evaluation}
\label{sec:sw-eval}



\subsection{Experimental Setting}

For in-depth evaluation, we took two popular BERT variations of Transformer models, RoBERTa \cite{liu2019roberta} and MobileBERT \cite{sun2020mobilebert}, fine-tuned for General Language Understanding Evaluation (GLUE, \cite{wang2018glue}) and SQuAD v1.1 ~\cite{rajpurkar2016squad}. For RoBERTa experiments, we used \emph{I-BERT}'s code-base\footnote{https://github.com/kssteven418/I-BERT} and reproduced the reported accuracy results on full-precision and INT8 models. For MobileBERT experiments, we adopted HuggingFace's implementation\footnote{https://huggingface.co/google/mobilebert-uncased} and followed the standard fine-tuning procedure. 

We trained \emph{NN-LUT} for the target functions (GELU, EXP, DIV, 1/SQRT) with appropriate input data and parameter initialization as discussed in Sec.~\ref{sec:strategy-param-init}. After the extensive empirical study, we concluded that a set of standard training hyper-parameters work well for all the non-linear operations tested in this work: learning-rate=0.001 (w/ multi-step), ADAM optimizer, and L1-Loss. We found that L1 loss slightly outperforms the other choices, partially due to modest penalization for the outliers. We also found that the dataset size of 100K was enough for curve-fitting. \emph{NN-LUT} training is straightforward and quick (it takes two minutes using one NVIDIA V100 GPU, and it is a one-time cost). 

Calibration can be further performed on the fine-tuned model. For demonstration, only one-tenth of the training dataset was used without labels, and the calibration repeats for five epochs, taking less than 5\% of the fine-tuning time. Note that approximation-aware fine-tuning (\cite{shkim2021ibert,stevens2021softermax}) can be orthogonally applied to adjust the original Transformer parameters and further compensate the accuracy gap if needed. Still, we found that the proposed calibration alone can boost the accuracy significantly.

Two other approximation methods are evaluated for performance comparison. We first constructed a linear-mode LUT (\emph{Linear-LUT}) by curve fitting with the 1st order polynomial. We also adopted \emph{I-BERT}'s state-of-the-art INT32 arithmetic approximation techniques. We used 16-entries for the LUT-based approximation with breakpoints and parameters represented in either FP32 or FP16 or INT32\footnote{FP16: convert FP32 values of breakpoints and parameters into FP16. INT32: adopt the scaling-factor calculation of \emph{I-BERT} to quantize FP32 values into INT32 directly}. From the ablation study, we found that 16-entries are enough for \emph{NN-LUT} to achieve high approximation accuracy.  

\subsection{Approximation Accuracy}

We first conducted an operation-wise evaluation of \emph{NN-LUT}. Figure~\ref{fig:expr-non-linear-ops} compares the approximation accuracy of \emph{NN-LUT} and \emph{Linear-LUT}. The top row illustrates the approximation results for GELU, Softmax, and LayerNorm on the selected input, and the bottom row shows L1 error. Note that both \emph{Linear-LUT} and \emph{NN-LUT} approximates GELU well thanks to its monotonous shape. In the case of Softmax and LayerNorm, however, points from \emph{Linear-LUT} have more deviation from FP32 points. These operations involve non-linear functions with a large dynamic range (e.g., division or 1/SQRT). Thus adjustable breakpoints of \emph{NN-LUT} improve approximation accuracy. This improved approximation capability of \emph{NN-LUT} becomes essential in successful Transformer inference.

\begin{table}[]

\begin{center}
\caption{Accuracy comparison for approximation of non-linear operations of RoBERTa on GLUE.} 
\label{tab:expr-glue}
\vspace{-3mm}
\resizebox{\linewidth}{!}{
{\footnotesize
\begin{tabular}{|ccccccccc|}
\multicolumn{9}{c}{(a) Direct approximation on FP32 RoBERTa pre-trained  model} \\
\hline
\multicolumn{1}{|c|}{Method}         & \multicolumn{1}{c|}{MRPC} & \multicolumn{1}{c|}{RTE}  & \multicolumn{1}{c|}{CoLA}  & \multicolumn{1}{c|}{SST-2} & \multicolumn{1}{c|}{STS-B} & \multicolumn{1}{c|}{QQP}  & \multicolumn{1}{c|}{MNLI} & QNLI \\ \hline
\multicolumn{1}{|c|}{Baseline}       & \multicolumn{1}{c|}{87.5} & \multicolumn{1}{c|}{79.4} & \multicolumn{1}{c|}{62.1}  & \multicolumn{1}{c|}{94.6}  & \multicolumn{1}{c|}{91.1}  & \multicolumn{1}{c|}{90.2} & \multicolumn{1}{c|}{87.9} & 92.8 \\ \hline
\multicolumn{9}{|c|}{\emph{Linear-LUT} (FP32)}                                                                                                                                                                                                            \\ \hline
\multicolumn{1}{|c|}{GELU only}      & \multicolumn{1}{c|}{87.8} & \multicolumn{1}{c|}{79.8} & \multicolumn{1}{c|}{62.1}  & \multicolumn{1}{c|}{94.5}  & \multicolumn{1}{c|}{91.1}  & \multicolumn{1}{c|}{90.2} & \multicolumn{1}{c|}{87.9} & 92.9 \\ \hline
\multicolumn{1}{|c|}{Softmax only}   & \multicolumn{1}{c|}{87.2} & \multicolumn{1}{c|}{78.3} & \multicolumn{1}{c|}{60.0}    & \multicolumn{1}{c|}{94.6}  & \multicolumn{1}{c|}{91.0}    & \multicolumn{1}{c|}{90.0}   & \multicolumn{1}{c|}{87.7} & 92.5 \\ \hline
\multicolumn{1}{|c|}{LayerNorm only} & \multicolumn{1}{c|}{57.5} & \multicolumn{1}{c|}{50.2} & \multicolumn{1}{c|}{4.6}   & \multicolumn{1}{c|}{80.0}    & \multicolumn{1}{c|}{35.7}  & \multicolumn{1}{c|}{46.8} & \multicolumn{1}{c|}{63.4} & 54.1 \\ \hline
\multicolumn{1}{|c|}{Altogether}     & \multicolumn{1}{c|}{60.2} & \multicolumn{1}{c|}{56.5} & \multicolumn{1}{c|}{4.8}   & \multicolumn{1}{c|}{82.6}  & \multicolumn{1}{c|}{30.2}  & \multicolumn{1}{c|}{45.9} & \multicolumn{1}{c|}{62.1} & 55.9 \\ \hline
\multicolumn{9}{|c|}{\emph{NN-LUT} (FP32)}                                                                                                                                                                                                                \\ \hline
\multicolumn{1}{|c|}{GELU only}      & \multicolumn{1}{c|}{87.5} & \multicolumn{1}{c|}{79.4} & \multicolumn{1}{c|}{61.7} & \multicolumn{1}{c|}{94.4}  & \multicolumn{1}{c|}{91.1}  & \multicolumn{1}{c|}{90.1} & \multicolumn{1}{c|}{87.9} & 92.8 \\ \hline
\multicolumn{1}{|c|}{Softmax only}   & \multicolumn{1}{c|}{87.5} & \multicolumn{1}{c|}{79.8} & \multicolumn{1}{c|}{61.6} & \multicolumn{1}{c|}{94.4}  & \multicolumn{1}{c|}{90.9}  & \multicolumn{1}{c|}{90.1} & \multicolumn{1}{c|}{88.0}   & 92.8 \\ \hline
\multicolumn{1}{|c|}{LayerNorm only} & \multicolumn{1}{c|}{86.5} & \multicolumn{1}{c|}{78.7} & \multicolumn{1}{c|}{60.4} & \multicolumn{1}{c|}{94.6}  & \multicolumn{1}{c|}{90.7}  & \multicolumn{1}{c|}{90.0}   & \multicolumn{1}{c|}{87.6} & 92.3 \\ \hline
\multicolumn{1}{|c|}{Altogether}     & \multicolumn{1}{c|}{87.4} & \multicolumn{1}{c|}{79.1} & \multicolumn{1}{c|}{61.7} & \multicolumn{1}{c|}{94.4}  & \multicolumn{1}{c|}{90.6}  & \multicolumn{1}{c|}{90.0}   & \multicolumn{1}{c|}{87.8} & 92.2 \\ \hline
\multicolumn{9}{c}{}                                                                                                                                                                                                                \\
\end{tabular}
}}

\resizebox{\linewidth}{!}{

\begin{tabular}{|ccccccccccc|}
\multicolumn{11}{c}{(b) INT8 RoBERTa pre-trained model (non-linear ops in FP32)} \\  
\hline

\multicolumn{1}{|c|}{Method}                  & \multicolumn{1}{c|}{Precision} & \multicolumn{1}{c|}{MRPC}          & \multicolumn{1}{c|}{RTE}           & \multicolumn{1}{c|}{CoLA}          & \multicolumn{1}{c|}{SST-2}         & \multicolumn{1}{c|}{STS-B}         & \multicolumn{1}{c|}{QQP}           & \multicolumn{1}{c|}{MNLI}        & \multicolumn{1}{c|}{QNLI}          & Avg           \\ \hline
\multicolumn{1}{|c|}{Baseline}                & \multicolumn{1}{c|}{FP32}      & \multicolumn{1}{c|}{88.7}          & \multicolumn{1}{c|}{77.3}          & \multicolumn{1}{c|}{61.6}          & \multicolumn{1}{c|}{94.6}          & \multicolumn{1}{c|}{91.0}            & \multicolumn{1}{c|}{90.2}          & \multicolumn{1}{c|}{87.5}        & \multicolumn{1}{c|}{92.6}          & 85.4          \\ \hline
\multicolumn{1}{|c|}{\emph{I-BERT}}                  & \multicolumn{1}{c|}{INT32}     & \multicolumn{1}{c|}{86.8}          & \multicolumn{1}{c|}{76.0}            & \multicolumn{1}{c|}{58.9}          & \multicolumn{1}{c|}{94.1}          & \multicolumn{1}{c|}{90.7}          & \multicolumn{1}{c|}{90.0}            & \multicolumn{1}{c|}{87.2}        & \multicolumn{1}{c|}{92.5}          & 84.5          \\ \hline
\multicolumn{1}{|c|}{\multirow{4}{*}{\emph{NN-LUT}}} & \multicolumn{1}{c|}{FP32}      & \multicolumn{1}{c|}{87.4}          & \multicolumn{1}{c|}{75.5}          & \multicolumn{1}{c|}{59.3}          & \multicolumn{1}{c|}{93.8}          & \multicolumn{1}{c|}{90.6}          & \multicolumn{1}{c|}{90.0}            & \multicolumn{1}{c|}{87.1}        & \multicolumn{1}{c|}{92.2}          & 84.5          \\ \cline{2-11} 
\multicolumn{1}{|c|}{}                        & \multicolumn{1}{c|}{FP32+C}    & \multicolumn{1}{c|}{\textbf{88.8}} & \multicolumn{1}{c|}{\textbf{75.5}} & \multicolumn{1}{c|}{\textbf{62.7}} & \multicolumn{1}{c|}{\textbf{93.7}} & \multicolumn{1}{c|}{\textbf{90.7}} & \multicolumn{1}{c|}{\textbf{89.5}} & \multicolumn{1}{c|}{\textbf{87.0}} & \multicolumn{1}{c|}{\textbf{92.6}} & \textbf{85.1} \\ \cline{2-11} 
\multicolumn{1}{|c|}{}                        & \multicolumn{1}{c|}{INT32}     & \multicolumn{1}{c|}{85.7}          & \multicolumn{1}{c|}{74.8}          & \multicolumn{1}{c|}{58.5}          & \multicolumn{1}{c|}{93.8}          & \multicolumn{1}{c|}{90.6}          & \multicolumn{1}{c|}{90.1}          & \multicolumn{1}{c|}{87.2}        & \multicolumn{1}{c|}{92.2}          & 84.1          \\ \cline{2-11} 
\multicolumn{1}{|c|}{}                        & \multicolumn{1}{c|}{INT32+C}   & \multicolumn{1}{c|}{86.8}          & \multicolumn{1}{c|}{78.0}          & \multicolumn{1}{c|}{61.4}          & \multicolumn{1}{c|}{94.0}            & \multicolumn{1}{c|}{90.6}          & \multicolumn{1}{c|}{90.0}            & \multicolumn{1}{c|}{87.1}        & \multicolumn{1}{c|}{92.5}          & 85.1          \\ \hline
\end{tabular}
}
\end{center}
\vspace{-5mm}
\end{table}

\begin{table}
\vspace{3mm}
\caption{Direct approximation of Softmax of MobileBERT on SQuAD.}
\label{tab:expr-squad}
\centering
\vspace{-3mm}
\resizebox{\linewidth}{!}{%
\begin{tabular}{|c|c|c|c|c|c|c|}
\hline
Approx. Type    & Baseline & \multicolumn{2}{c|}{\emph{Linear-LUT}} & \multicolumn{2}{c|}{\emph{NN-LUT}} \\ \hline
Softmax Prec & FP32             & FP32               & FP16              & FP32             & FP16            \\ \hline
F1 score (loss) & 89.3             & 87.8 (-1.5)        & 87.7 (-1.6)       & 89.3 (0.0)       & 89.3 (0.0)      \\ \hline
\end{tabular}%
}
\vspace{-5mm}
\end{table}

\subsection{Transformer Inference Accuracy}

For an in-depth evaluation of \emph{NN-LUT}'s approximation capability, we conducted BERT inference with the non-linear operations approximated in various settings. Table~\ref{tab:expr-glue} summarizes the accuracy comparison on GLUE tasks. We first compare the performance of \emph{Linear-LUT} and \emph{NN-LUT} on the full-precision RoBERTa baseline. Input scaling is applied to both methods for LayerNorm. Note that this is \textit{direct approximation}; we do not perform time-consuming fine-tuning to compensate for approximation error. Both LUTs are implemented in FP32. As shown in Table~\ref{tab:expr-glue}(a), \emph{Linear-LUT} suffers significant accuracy loss while \emph{NN-LUT} achieves almost the same inference accuracy compared to the FP32. LayerNorm is the most sensitive to the approximation, especially for \emph{Linear-LUT}; input scaling does not help due to its fixed breakpoints.

Next, we compare approximation performance with the state-of-the-art technique, \emph{I-BERT}. The experiments adopted \emph{I-BERT}'s \textit{reduced-precision} RoBERTa code-base; as a baseline, the model is fine-tuned with INT8 matrix multiplication and FP32 non-linear operations. Table~\ref{tab:expr-glue}(b) summarizes the accuracy comparison between \emph{I-BERT} and \emph{NN-LUT}. In addition to FP32 implementation, we constructed INT32 \emph{NN-LUT} to match the INT32 precision setting of \emph{I-BERT}'s non-linear operations. As shown in the table, the accuracy of FP32 \emph{NN-LUT} is on-par with \emph{I-BERT}, and INT32 \emph{NN-LUT} further experiences slight accuracy degradation. 

A notable advantage of \emph{NN-LUT} is that it can be calibrated to decrease the approximation errors further. Noting that LayerNorm is the most sensitive non-linear operation (from Table~\ref{tab:expr-glue}(a)), we conducted a calibration only for \emph{NN-LUT} on the LayerNorm operations. As shown in Table~\ref{tab:expr-glue}(b), average accuracy is significantly increased for both FP32 and INT32 \emph{NN-LUT}, surpassing the accuracy of \emph{I-BERT}. This accuracy boost by calibration differentiates NN-LUT from conventional approximation methods.

To further demonstrate the superior performance of \emph{NN-LUT} on a different NLP task, model, and precision setting, we perform direct approximation on MobileBERT for the question answering task (SQuAD v1.1). Note that Softmax is the only non-linear operation involved in the transformer layer of MobileBERT. Table~\ref{tab:expr-squad} shows the accuracy comparison between \emph{Linear-LUT} and \emph{NN-LUT} implemented in FP32 and FP16. In all the cases, MatMul is computed in FP16. Similar to the previous experiments, \emph{NN-LUT} achieves the baseline accuracy while \emph{Linear-LUT} suffers noticeable accuracy degradation in both precision settings.  


\begin{figure*}[!t]
\centering
\includegraphics[width=0.8\linewidth]{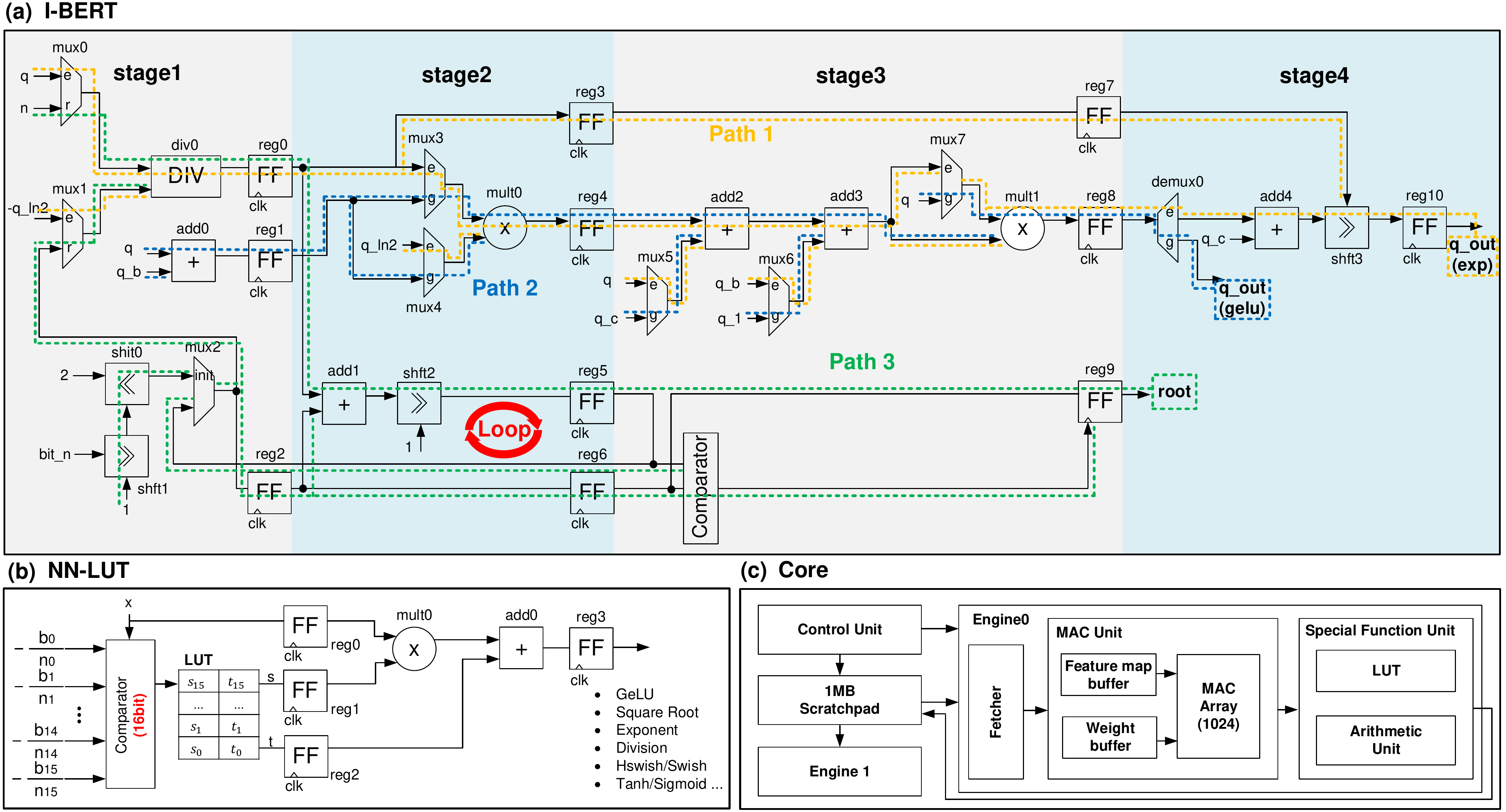}
\vspace{-5mm}
\caption{Architecture overview: (a) \emph{NN-LUT}, (b) \emph{I-BERT} (based on approximation algorithms of \cite{shkim2021ibert}), (c) accelerator core.}
\label{fig:hw-arch}
\vspace{-3mm}
\end{figure*}

\begin{table}[]

\caption{Performance comparison of arithmetic units for approximation of non-linear operations.}
\vspace{-3mm}
\label{tab:expr-hw-perf}
\centering
\resizebox{\linewidth}{!}{%
\begin{tabular}{|c|ccc|ccc|}
\hline
Approximation                    & \multicolumn{3}{c|}{\emph{I-BERT}}                                       & \multicolumn{3}{c|}{\emph{NN-LUT} (16-entry)}                                  \\ \hline
Precision                        & \multicolumn{3}{c|}{INT32}                                        & \multicolumn{1}{c|}{INT32}   & \multicolumn{1}{c|}{FP16}     & FP32     \\ \hline
Area (um2)                       & \multicolumn{3}{c|}{2654.32}                                     & \multicolumn{1}{c|}{1008.92} & \multicolumn{1}{c|}{498.38} & 1133.60 \\ \hline
Power (mW)                       & \multicolumn{3}{c|}{2.1421}                                       & \multicolumn{1}{c|}{0.0591}  & \multicolumn{1}{c|}{0.0250}    & 0.0437   \\ \hline
Delay (ns)                       & \multicolumn{3}{c|}{2.67}                                         & \multicolumn{1}{c|}{0.68}    & \multicolumn{1}{c|}{1.36}     & 1.60      \\ \hline
\multirow{2}{*}{Latency (cycle)} & \multicolumn{1}{c|}{I-GELU} & \multicolumn{1}{c|}{I-EXP} & I-SQRT & \multicolumn{3}{c|}{GELU, EXP, DIV, 1/SQRT}                             \\ \cline{2-7} 
                                 & \multicolumn{1}{c|}{3}      & \multicolumn{1}{c|}{4}     & 5      & \multicolumn{3}{c|}{2}                                                  \\ \hline

\end{tabular}%
}
\vspace{-5mm}
\end{table}

\section{Hardware Evaluation}
\label{sec:hw-eval}

The previous evaluation of approximation performance has demonstrated that \emph{NN-LUT} works as an accurate and general approximator for non-linear operations of Transformers. In this section, we evaluate hardware advantages of \emph{NN-LUT} by implementing custom hardware and conducting system-level performance analysis.

\subsection{Comparison of Approximation Hardware}

We carefully implemented the arithmetic units for \emph{NN-LUT} and \emph{I-BERT} and compared their hardware cost. Following the approach of \cite{shkim2021ibert}'s implementation, we assume that the input is pre-scaled to the target bit-precision\footnote{In \emph{I-BERT} implementation at https://github.com/kssteven418/I-BERT, the integer input of non-linear operations is pre-scaled with scale-factors as the output of the previous computation layer.}. The breakpoints $d$, the approximation parameters $s,t$, and the associated computing units follow the input's bit-precision.    

We design the arithmetic units covering the required sequence of computation for approximation of non-linear operations. The differences of the designed arithmetic units between \emph{NN-LUT} and \emph{I-BERT} are summarized as follows (Figure~\ref{fig:hw-arch}(a,b):

\begin{itemize}
    \item \textbf{\emph{NN-LUT}}: A comparator for index checking and a LUT for look-up of the approximation parameters ($a_i, b_i$) for each input element, followed by a multiplier and an adder for computation of approximated output.
    \item \textbf{\emph{I-BERT}}: Integration of multipliers, adders, shifters, and a divider to execute the computation sequences of \emph{I-BERT}'s integer-only GELU, EXP, and SQRT (Algorithm 2-4 of \cite{shkim2021ibert}).  
\end{itemize}

We synthesized these arithmetic units using the 7-nm technology and obtained the area, power consumption, and critical path delay. Also, we measure latency as the number of cycles for executing non-linear functions such as GELU, EXP, DIV, and SQRT in each arithmetic unit. Table~\ref{tab:expr-hw-perf} summarizes the hardware performance comparison. As shown in the table, \emph{I-BERT} suffers significantly higher area, power consumption, and delay compared to \emph{NN-LUT}. In case of INT32 implementation, \emph{I-BERT} takes up $2.63\times$ area, $36.4\times$ power consumption, and $3.93\times$ delay compared to \emph{NN-LUT}. Furthermore, \emph{I-BERT} takes long latency due to its repetitive computation sequences of \emph{I-BERT}, following different data paths of \emph{I-BERT} arithmetic unit. In contrast, LUT-based approximation takes two cycles for look-up and computation for all the non-linear operations. 
Therefore, by replacing non-linear operations with \emph{NN-LUT}, we can expect a significant performance boost while maintaining inference accuracy.

\subsection{Accelerator Integration}
To understand the hardware benefit of \emph{NN-LUT} for Transformer inference, we perform a system-level integration of different approximation hardware with a neural processing unit. 
As a concrete example, we construct a cycle-accurate hardware simulator inspired by the mobile DNN accelerator architecture \cite{song20197,jang2021sparsity}. As shown in Figure~\ref{fig:hw-arch}(c), the accelerator core consists of a control unit, a shared scratch-pad, two compute engines and a special function unit. Each engine is equipped with a 32x32 MAC array capable of 64 dot-products of 16-dimensional vectors every cycle, producing a partial-sum vector of 16 output channels. It is followed by a vector of special function units for the throughput matching calculation of activation functions. Since the special function unit is already equipped with LUT and the arithmetic unit, as does in many existing NPUs (\cite{NVDLA,jang2021sparsity}), implementation of \emph{NN-LUT} might incur little additional hardware cost.


\subsection{System-Level Performance Analysis}
Table~\ref{tab:sys-perf} shows the breakdown of relative cycles of RoBERTa inference computation with the increasing sequence length (SL). The observations are summarized as follows:

\begin{itemize}
    \item In the case of \emph{I-BERT}, the execution cycles corresponding to non-linear operations is significant, growing from 17.7\% (SL=16) to 37.8\% (SL=1024). These results concur with the previous observations that non-linear operations become serious overhead \cite{sun2020mobilebert,stevens2021softermax}.
    \item In the case of \emph{NN-LUT}, the portion for non-linear operations is significantly reduced (up to 43\% at SL=1024), demonstrating the superior efficiency of \emph{NN-LUT} compared to \emph{I-BERT}.
    \item Thanks to \emph{NN-LUT}'s performance improvement, there can be up to 26\% speedup in total execution time compared to \emph{I-BERT}.
\end{itemize}
From this detailed performance analysis, we could conclude that \emph{NN-LUT} provides non-trivial benefits in Transformer computation.


\begin{table}[]
\caption{System-level performance comparison.}
\vspace{-3mm}
\label{tab:sys-perf}
\centering
\resizebox{\linewidth}{!}{%
\begin{tabular}{|cc|cccccccc|}
\hline
\multicolumn{2}{|c|}{RoBERTa}                                  & \multicolumn{8}{c|}{Relative computation cycles (\%)}                                                                                                                                                            \\ \hline
\multicolumn{1}{|c|}{Operations}                  & Seq-Length & \multicolumn{1}{c|}{16}    & \multicolumn{1}{c|}{32}    & \multicolumn{1}{c|}{64}    & \multicolumn{1}{c|}{128}   & \multicolumn{1}{c|}{256}   & \multicolumn{1}{c|}{384}   & \multicolumn{1}{c|}{512}   & 1024  \\ \hline\hline
\multicolumn{1}{|c|}{\multirow{3}{*}{\emph{I-BERT} Ops}} & GELU       & \multicolumn{1}{c|}{6.55}  & \multicolumn{1}{c|}{6.58}  & \multicolumn{1}{c|}{6.45}  & \multicolumn{1}{c|}{6.22}  & \multicolumn{1}{c|}{5.80}  & \multicolumn{1}{c|}{5.43}  & \multicolumn{1}{c|}{5.11}  & 4.12  \\ \cline{2-10} 
\multicolumn{1}{|c|}{}                            & LayerNorm  & \multicolumn{1}{c|}{9.82}  & \multicolumn{1}{c|}{9.86}  & \multicolumn{1}{c|}{9.68}  & \multicolumn{1}{c|}{9.33}  & \multicolumn{1}{c|}{8.70}  & \multicolumn{1}{c|}{8.14}  & \multicolumn{1}{c|}{7.66}  & 6.19  \\ \cline{2-10} 
\multicolumn{1}{|c|}{}                            & Softmax    & \multicolumn{1}{c|}{1.36}  & \multicolumn{1}{c|}{1.37}  & \multicolumn{1}{c|}{2.69}  & \multicolumn{1}{c|}{5.18}  & \multicolumn{1}{c|}{9.66}  & \multicolumn{1}{c|}{13.57} & \multicolumn{1}{c|}{17.02} & 27.49 \\ \hline
\multicolumn{2}{|c|}{MatMul}                                   & \multicolumn{1}{c|}{81.17} & \multicolumn{1}{c|}{81.64} & \multicolumn{1}{c|}{80.65} & \multicolumn{1}{c|}{78.76} & \multicolumn{1}{c|}{75.36} & \multicolumn{1}{c|}{72.40} & \multicolumn{1}{c|}{69.79} & 61.86 \\ \hline
\multicolumn{2}{|c|}{etc.}                                     & \multicolumn{1}{c|}{1.09}  & \multicolumn{1}{c|}{0.55}  & \multicolumn{1}{c|}{0.54}  & \multicolumn{1}{c|}{0.52}  & \multicolumn{1}{c|}{0.48}  & \multicolumn{1}{c|}{0.45}  & \multicolumn{1}{c|}{0.43}  & 0.34  \\ \hline\hline
\multicolumn{1}{|c|}{\multirow{3}{*}{\emph{NN-LUT} Ops}} & GELU       & \multicolumn{1}{c|}{4.71}  & \multicolumn{1}{c|}{4.73}  & \multicolumn{1}{c|}{4.68}  & \multicolumn{1}{c|}{4.57}  & \multicolumn{1}{c|}{4.37}  & \multicolumn{1}{c|}{4.19}  & \multicolumn{1}{c|}{4.02}  & 3.46  \\ \cline{2-10} 
\multicolumn{1}{|c|}{}                            & LayerNorm  & \multicolumn{1}{c|}{5.89}  & \multicolumn{1}{c|}{5.92}  & \multicolumn{1}{c|}{5.85}  & \multicolumn{1}{c|}{5.71}  & \multicolumn{1}{c|}{5.46}  & \multicolumn{1}{c|}{5.24}  & \multicolumn{1}{c|}{5.03}  & 4.33  \\ \cline{2-10} 
\multicolumn{1}{|c|}{}                            & Softmax    & \multicolumn{1}{c|}{0.59}  & \multicolumn{1}{c|}{0.59}  & \multicolumn{1}{c|}{1.17}  & \multicolumn{1}{c|}{2.29}  & \multicolumn{1}{c|}{4.37}  & \multicolumn{1}{c|}{6.28}  & \multicolumn{1}{c|}{8.04}  & 13.85 \\ \hline
\multicolumn{2}{|c|}{MatMul}                                   & \multicolumn{1}{c|}{87.63} & \multicolumn{1}{c|}{88.17} & \multicolumn{1}{c|}{87.72} & \multicolumn{1}{c|}{86.86} & \multicolumn{1}{c|}{85.25} & \multicolumn{1}{c|}{83.77} & \multicolumn{1}{c|}{82.41} & 77.92 \\ \hline
\multicolumn{2}{|c|}{etc.}                                     & \multicolumn{1}{c|}{1.18}  & \multicolumn{1}{c|}{0.59}  & \multicolumn{1}{c|}{0.58}  & \multicolumn{1}{c|}{0.57}  & \multicolumn{1}{c|}{0.55}  & \multicolumn{1}{c|}{0.52}  & \multicolumn{1}{c|}{0.50}  & 0.43  \\ \hline\hline
\multicolumn{2}{|c|}{Speedup (times)}                          & \multicolumn{1}{c|}{1.08}  & \multicolumn{1}{c|}{1.08}  & \multicolumn{1}{c|}{1.09}  & \multicolumn{1}{c|}{1.10}  & \multicolumn{1}{c|}{1.13}  & \multicolumn{1}{c|}{1.16}  & \multicolumn{1}{c|}{1.18}  & 1.26  \\ \hline
\end{tabular}
}
\vspace{-3mm}
\end{table}

\section{Conclusion}
\label{sec:conclusion}

We propose a novel neural network approximation method for non-linear operations of pre-trained Transformers. We show that a trained approximation network can be converted to a look-up table implementation with equivalent approximation functionality. The experimental results show that the proposed method, called \emph{NN-LUT} can work as a drop-in replacement of non-linear operations in RoBERTa and MobileBERT without accuracy degradation while improving hardware performance by $2.63\times$ in area, $36.4\times$ in power consumption, and $3.93\times$ in delay. Furthermore, based on the system-level performance analysis, \emph{NN-LUT} achieves up to 26\% speedup compared to the state-of-the-art alternative.


\bibliographystyle{ACM-Reference-Format-ks}
\bibliography{reference}

\appendix

\end{document}